\documentclass[journal, 10pt, twoside]{IEEEtran}

\usepackage{amssymb}
\usepackage[ruled,linesnumbered]{algorithm2e}
\usepackage{cuted}
\usepackage{balance}
\usepackage[hidelinks=true,bookmarks=FWLse]{hyperref}
\usepackage{lineno}
\usepackage{amsopn}
\usepackage{comment}
\usepackage{color}
\usepackage{cite}
\usepackage{algorithmic}
\usepackage{xcolor}
\usepackage{nomencl}
\usepackage{enumitem}

\usepackage[T1]{fontenc}
\usepackage{amsmath}
\usepackage{subfigure} 

\usepackage{graphicx}
\usepackage{color}
\usepackage{tabulary}
\usepackage{booktabs}
\usepackage{multirow}
\usepackage{makecell}
\usepackage{supertabular}
\usepackage[linesnumbered,ruled]{algorithm2e}
\usepackage{url}
\DeclareMathOperator*{\argminA}{arg\,min}
\newcommand\norm[1]{\left\lVert#1\right\rVert}

\usepackage{nomencl}

\makenomenclature

\newcommand{\ie}{\textit{i.e.}}
\newcommand{\eg}{\textit{e.g.}}

\newcommand{\etal}{\textit{et al.}}

\title{Rethinking Road Surface 3D Reconstruction and  Pothole Detection: From Perspective Transformation to Disparity Map Segmentation}
\author{Rui~Fan,~\IEEEmembership{Member,~IEEE}, Umar~Ozgunalp,~\IEEEmembership{Member,~IEEE},\\ Yuan~Wang,~Ming~Liu,~\IEEEmembership{Senior~Member,~IEEE,}
Ioannis~Pitas,~\IEEEmembership{Fellow,~IEEE
}
\thanks{R. Fan is with UC San Diego. e-mail:  rui.fan@ieee.org}
\thanks{U. Ozgunalp is with the Department of Electrical and Electronics Engineering,
	Cyprus International University, Mersin, Turkey. e-mail: uozgunalp@ciu.edu.tr}
\thanks{Y. Wang and M. Liu are  with the Department of Electronic and Computer Engineering,  the Hong Kong University of Science and Technology, Hong Kong. e-mail:  \{ywangeq, eelium\}@ust.hk}
\thanks{I. Pitas is with the Department of Informatics, University of Thessaloniki, Greece. e-mail: pitas@aiia.csd.auth.gr
}
}

\begin{document}
\maketitle

\begin{abstract}
Potholes are one of the most common forms of road damage, which can severely affect driving comfort, road safety and vehicle condition. Pothole detection is typically performed by either structural engineers or certified inspectors. This task is, however, not only hazardous for the personnel but also extremely time-consuming. This paper presents an efficient pothole detection algorithm based on road disparity map estimation and segmentation. We first generalize the perspective transformation by incorporating the stereo rig roll angle. The road disparities are then estimated using semi-global matching. A disparity map transformation algorithm is then performed to better distinguish the damaged road areas. Finally, we utilize simple linear iterative clustering to group the transformed disparities into a collection of superpixels. The potholes are then detected by finding the superpixels, whose values are lower than an adaptively determined threshold.
The proposed algorithm is implemented  on an NVIDIA RTX 2080 Ti GPU in CUDA. The experiments demonstrate the accuracy and efficiency of our proposed road pothole detection algorithm, where an accuracy of $\textbf{99.6\%}$ and an F-score of $\textbf{89.4\%}$ are achieved.

\end{abstract}

\IEEEpeerreviewmaketitle



\section{Introduction}
\label{sec.pt_introduction}

\IEEEPARstart{A} pothole is a considerably large structural road failure, caused by the contraction and expansion of rainwater that permeates into the ground, under the road surface \cite{Miller2014}. Frequently inspecting and repairing potholes is crucial for road maintenance \cite{fan2018thesis}. Potholes are regularly detected and reported by certified inspectors and structural engineers \cite{Kim2014}. However, this process is not only time-consuming and costly, but also dangerous for the personnel \cite{Fan2018}. Additionally, such a detection is always qualitative and subjective, because decisions depend  entirely on the individual experience \cite{Koch2015}. Therefore, there is an ever-increasing need to develop a robust and precise automated road condition assessment system that can detect potholes both quantitatively and objectively. \cite{Mathavan2015}.

Over the past decade, various technologies, such as vibration sensing, active sensing and passive sensing, have been utilized to acquire road data and detect road damages. For example, Fox {\etal} \cite{Fox2017} developed a crowd-sourcing system to detect and localize potholes by analyzing the accelerometer data obtained from multiple vehicles. Vibration sensors are cost-effective and only require a small storage space. However, the pothole shape and volume cannot be explicitly inferred from the vibration sensor data \cite{Fan2018}. Additionally, road hinges and joints are often mistaken for potholes \cite{Kim2014}. Therefore,  researchers have been focused towards developing pothole detection systems based on active and passive sensing. For instance, Tsai and Chatterjee \cite{Tsai2017} mounted two laser scanners on a Georgia Institute of Technology Sensing Vehicle (GTSV) to collect 3D road data for pothole detection. However, such vehicles are not widely used, because of high equipment purchase and long-term maintenance costs \cite{Fan2018}.


The most commonly used passive sensors for pothole detection include Microsoft Kinect and other types of digital cameras \cite{Wang2017}. For example, a Kinect was used to acquire road depth information, and  image segmentation algorithms were applied for pothole detection \cite{Jahanshahi2012}. However, Kinect was not designed for outdoor use, and often fails to perform well, when exposed to direct sunlight, resulting in wrong (zero) depth values \cite{Jahanshahi2012}. Therefore, it is more effective to detect potholes using digital cameras, as they are cost-effective and capable of working in outdoor environments \cite{Fan2018}. Given the dimensions of the acquired road data, passive sensing (computer vision) approaches \cite{Wang2017} are generally grouped into two categories: 2D vision-based and 3D reconstruction-based ones \cite{Koch2012} . 

The 2D vision-based road pothole detection algorithms generally comprise three steps: image segmentation, contour extraction and object recognition \cite{Kim2014}. These methods are usually developed based on the following hypotheses \cite{Koch2012}: 
\begin{itemize}
	\item potholes are concave holes;
	\item the pothole texture  is grainier and coarser than that of the surrounding road surface; and 
	\item the intensities of the pothole region-of-interest (RoI) pixels are typically lower than those of the surrounding road surface, due to shadows. 
\end{itemize}

\begin{figure*}[!t]
	\begin{center}
		\centering
		\includegraphics[width=0.82\textwidth]{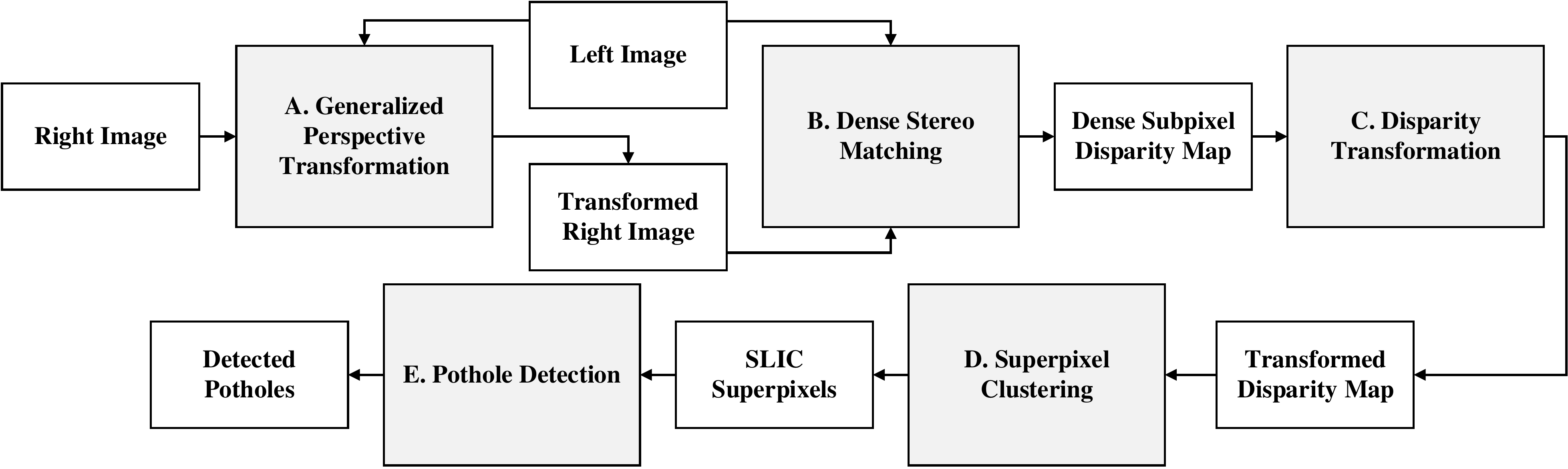}
		\centering
		\caption{The block diagram of our proposed road pothole detection system.}
		\label{fig.block_diagram}
	\end{center}
\end{figure*}

Basic image segmentation algorithms are first applied on  RGB or gray-scale road surface images to separate the damaged and undamaged road areas. Most commonly used segmentation algorithms are triangle thresholding \cite{Koch2011} and Otsu's thresholding \cite{Buza2013}. Compared with the former one, Otsu's thresholding algorithm minimizes the intra-class variance and exhibits better damaged road region detection accuracy \cite{Pitas2000}. Next, image filtering \cite{Tedeschi2017}, edge detection \cite{Ryu2015}, region growing \cite{Ouma2017} and morphological operations \cite{Li2016} are utilized to reduce redundant information (typically noise) and clarify the potential pothole RoI contour \cite{Koch2015}. The resulting pothole RoI is then modeled by an ellipse \cite{Koch2012, Miller2014, Tedeschi2017, Tsai2017}. Finally, the image texture within this  elliptical region  is compared with that of the surrounding road region. If the elliptical RoI has coarser and grainier texture than that of the surrounding region, a pothole is detected \cite{Koch2012}.

Although such 2D computer vision methods can recognize road  potholes with low computational complexity, the achieved detection and localization accuracy is still far from satisfactory \cite{Koch2012, Jahanshahi2012}. Furthermore, since the actual pothole contour is always irregular, the geometric assumptions made in contour extraction step can be ineffective. In addition, visual environment variability, such as road image texture, also significantly affects  segmentation results  \cite{Salari2011}. Therefore, machine learning methods \cite{Cord2012, Bray2006, Li2011} have been employed for better road pothole detection accuracy. For example,  Adaptive Boost (AdaBoost) was utilized to determine whether or not a road image contains damaged road RoI \cite{Cord2012}. Bray {\etal} \cite{Bray2006} also trained a neural network (NN) to detect and classify road damage. However, supervised classifiers require a large amount of labeled training data. Such data labeling procedures  can be very labor-intensive \cite{Koch2015}. 
Additionally, the 3D pothole model cannot be obtained by using only a single image. Therefore,  the depth information has proven to be more effective than the RGB information for detecting gross road damages, \eg, potholes \cite{Mathavan2015}. Therefore, the main purpose of this paper is to present a novel road pothole detection algorithm based on its 3D geometry reconstruction.
Multiple (at least two) camera views are required to this end \cite{Hartley2003}. Images from different viewpoints can be captured using either a single moving camera or a set of synchronized multi-view cameras \cite{Fan2018}. In \cite{Jog2012}, a single camera was mounted at the rear of the car to capture visual 2D road footage. Then, Scale-Invariant Feature Transform (SIFT) \cite{Lowe2004} feature points were extracted in each video frame, respectively. The matched SIFT feature correspondences on two consecutive video frames can be used to find the fundamental matrix. Then, a cost energy  related to all fundamental matrices was minimized using bundle adjustment. Each camera pose was, therefore, refined and the 3D geometry can be reconstructed in a Structure from Motion (SfM) manner \cite{Jog2012,Hartley2003}. However, SfM can only acquire sparse point clouds, which renders  pothole detection infeasible. Therefore, pothole detection using stereo vision technology has been researched in recent years, as it can provide dense disparity maps \cite{Fan2018}.

The first reported effort on employing stereo vision  for road damage detection utilized a camera pair and a structured light projector to acquire 3D crack and pothole models \cite{Barsi2005}. In recent years, surface modeling (SM)  has become a popular and effective technique for pothole detection \cite{Zhang2014, Ozgunalp2016, Mikhailiuk2016}. For example, in \cite{Zhang2014},  the road surface point cloud was represented by a quadratic model. Then, pothole  detection was straightforwardly realized by finding the points whose height is lower than the one of the modeled road surface. In \cite{Ozgunalp2016}, this approach was improved by adding a smoothness term to the residual function that is related to the planar patch orientation. This greatly minimizes the outlier effects caused by obstacles and can, therefore, provide more precise road surface modeling results. However, finding the best value for the smoothness term is a challenging task, as it may vary from case to case \cite{Mikhailiuk2016}. Similarly, Random Sample Consensus (RANSAC) was utilized to reduce outlier effects, while fitting a quadratic surface model to a disparity map rather than a point cloud \cite{Mikhailiuk2016}. This helps the RANSAC-SM algorithm perform more accurately and efficiently than both  \cite{Zhang2014} and \cite{Ozgunalp2016}. 

Road surface modeling and pothole detection are still open research. One problem is that the actual road surface is  sometimes uneven, which renders quadratic surface modeling somewhat problematic. Moreover, although comprehensive studies of 2D and 3D computer vision techniques for pothole detection have been made, these two categories are usually implemented independently \cite{Koch2015}. Their combination can possibly advance current state of the art to achieve highly accurate pothole detection results. For instance, our recent work \cite{fan2019pothole} combined both iterative disparity transformation and  3D road surface  modeling together for pothole detection. Although \cite{fan2019pothole} is computationally intensive, its achieved successful detection rate and overall pixel-level accuracy is much higher than both  \cite{Zhang2014} and \cite{Mikhailiuk2016}.

Therefore, in this paper, we present an efficient and robust pothole detection algorithm based on
road disparity map estimation and segmentation. The block diagram  of our proposed pothole detection algorithm is shown in Fig. \ref{fig.block_diagram}. Firstly, we generalize the perspective transformation (PT) proposed in \cite{Fan2018}, by incorporating the stereo rig roll angle into the PT process, which not only increases disparity estimation accuracy but also reduces its computational complexity \cite{Fan2018}. Due to its inherent parallel efficiency, semi-global matching (SGM) \cite{Hirschmuller2008} is utilized for  dense subpixel disparity map estimation. 
A fast disparity transformation (DT) algorithm is then performed on the estimated subpixel disparity maps to better distinguish between damaged and undamaged road regions, whereby minimizing an energy function with respect to the stereo rig roll angle and the road disparity projection model. 
Finally, we use simple linear iterative clustering (SLIC) algorithm \cite{Achanta2012} to group the transformed disparities into a collection of superpixels. The potholes are subsequently detected by finding the superpixels, whose values are lower than an adaptively determined threshold. Different potholes are also labeled using connect component labeling (CCL) \cite{Pitas2000}.

The remainder of this paper continues in the following manner: Section \ref{sec.algorithm_description} introduces the proposed pothole detection system. Section \ref{sec.exp} presents the experimental results and discusses the performance of the proposed system. In Section \ref{sec.discussion}, we discuss the practical application of our system. Finally, Section \ref{sec.con} concludes the paper and provides recommendations for future work. 

\section{Algorithm Description}
\label{sec.algorithm_description}

\subsection{Generalized Perspective Transformation}
\label{sec.PT}
Pothole detection focuses entirely on the road surface, which can be treated as a ground plane. Referring to the u-v-disparity analysis provided in \cite{hu2005complete}, when the stereo rig roll angle $\phi$ is neglected, the road disparity projections in the v-disparity domain can be represented by a straight line: $f(\mathbf{a}, \mathbf{p})=a_0+a_1 v$, where $\mathbf{a}=[a_0,a_1]^\top$ and $\mathbf{p}=[u,v]^\top$ is a pixel in the disparity map. $\mathbf{a}$ can be obtained by minimizing \cite{fan2020learning}: 
\begin{equation}
\begin{split}
E_0=\norm{\mathbf{d}-\Big[\mathbf{1}_k\ \  \mathbf{v}\Big]\mathbf{a}}^2_2,
\end{split}
\label{eq.E0_1}
\end{equation}
where 
$\mathbf{d}=[d_1,\cdots, d_k]^\top$ is a $k$-entry vector of road  disparity values, $\mathbf{1}_k$ is a $k$-entry vector of ones, and  $\mathbf{v}=[v_1,\cdots,v_k]^\top$ is a $k$-entry vector of the vertical coordinates of the road disparities.
However, in practice, $\phi$ is always non-zero, resulting in the disparity map to be rotated by $\phi$ around the image center. This leads to gradual disparity change in the horizontal direction, as shown in Fig. \ref{fig.disp1}. Applying an inverse rotation by $\phi$ on the original disparity map yields \cite{fan2018novel}:
\begin{equation}
\mathbf{p}' = \begin{bmatrix}
\cos\phi & \sin\phi \\
-\sin\phi & \cos\phi
\end{bmatrix}\mathbf{p},
\end{equation}
where $\mathbf{p}'$ represents the correspondence of $\mathbf{p}$ in the rotated disparity map. Therefore, the road disparity projections in the v-disparity domain can be represented by:  
\begin{equation}
f(\mathbf{a},\mathbf{p},\phi)=a_0+a_1(v\cos\phi-u\sin\phi). 
\label{eq.pt}
\end{equation}
Compared to \cite{Fan2018}, (\ref{eq.pt}) depicts PT in a more general way, as both $\mathbf{a}$ and $\phi$ are considered. The estimation of $\mathbf{a}$ and $\phi$ will be discussed in Section \ref{sec.disparity_transformation}. (\ref{eq.E0_1}) can, therefore, be rewritten as follows:
\begin{equation}
\begin{split}
E_0=\big|\big|\mathbf{d}-\mathbf{T}(\phi)\mathbf{a}\big|\big|^2_2,
\end{split}
\label{eq.E0}
\end{equation}
where 
\begin{equation}
\mathbf{T}(\phi)=\Big[\mathbf{1}_k \ \ \cos\phi\mathbf{v}-\sin\phi\mathbf{u}\Big],
\end{equation}
$\mathbf{u}=[u_1,\dots,u_k]^\top$ is a $k$-entry vector of the horizontal coordinates of the road disparities. Similar to \cite{Fan2018}, the PT can be realized by shifting each pixel on row $v$ in the right image $\kappa(\mathbf{a}, \mathbf{p}, \phi)$ pixels to the left, where $\kappa$ can be computed using:
\begin{equation}
\kappa(\mathbf{a}, \mathbf{p}, \phi)=\min_{x=0}^{W} \Big[a_0+a_1(v\cos\phi-x\sin\phi)-\delta_\text{PT}\Big],
\end{equation}
where $W$ denotes the maximum horizontal coordinate and $\delta_\text{PT}$ is a constant set to ensure that the values in the disparity map with respect to the original left image (see Fig. \ref{fig.left}) and the transformed right image (see Fig. \ref{fig.right}) are non-negative.

\subsection{Dense Stereo Matching}
\label{eq.disparity_estimation}
\begin{figure}[!t]
	\begin{center}
		\centering
		\subfigure[]
		{
			\includegraphics[width=0.20\textwidth]{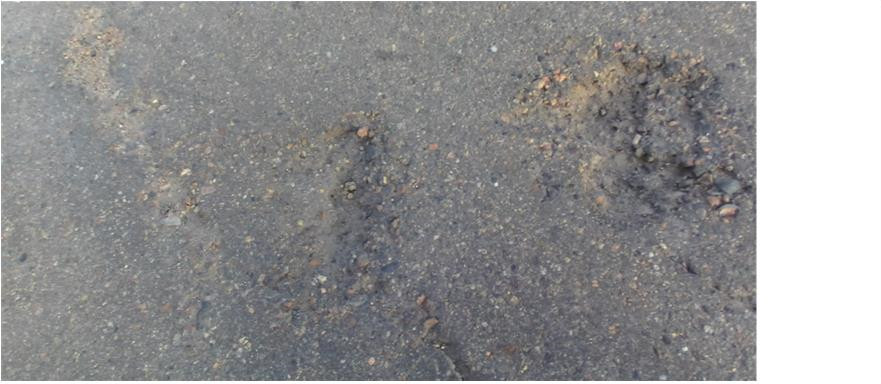}
			\label{fig.left}
		}
		\subfigure[]
		{
			\includegraphics[width=0.20\textwidth]{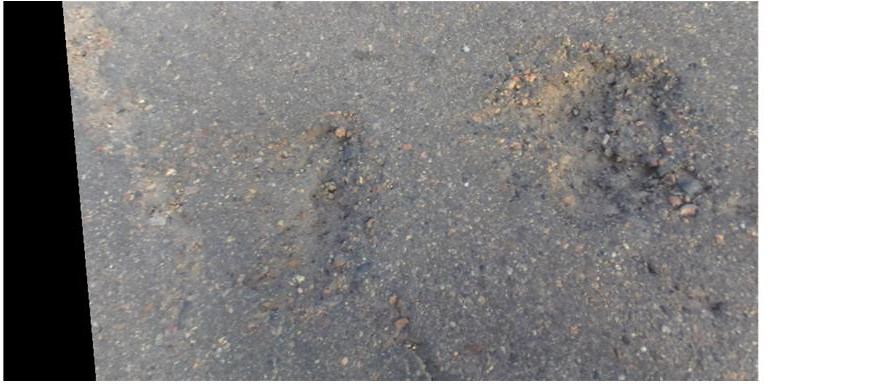}
			\label{fig.right}
		}\\
		\subfigure[]
		{
			\includegraphics[width=0.20\textwidth]{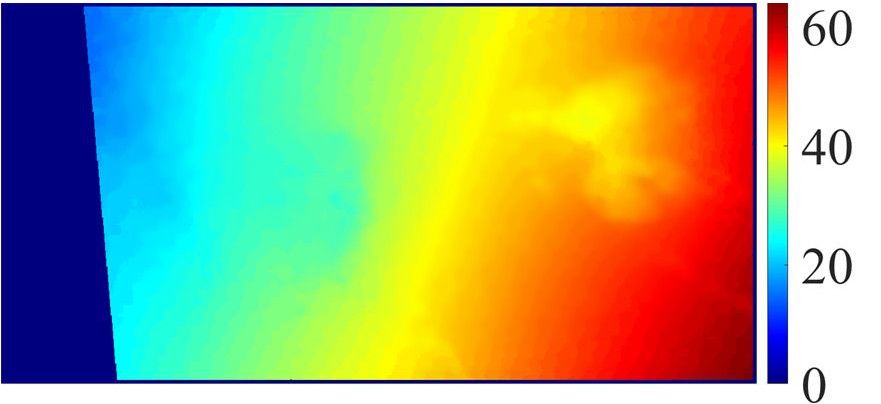}
			\label{fig.disp0}
		}
		\subfigure[]
		{
			\includegraphics[width=0.20\textwidth]{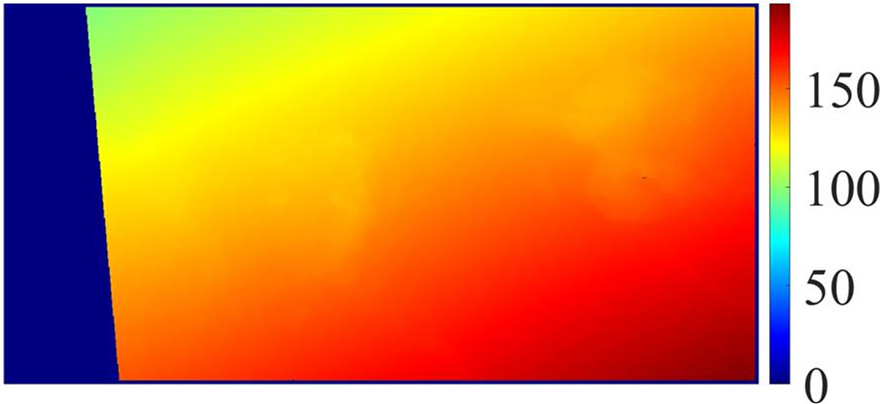}
			\label{fig.disp1}
		}
		\caption{Generalized perspective transformation and disparity estimation:  (a) original left image; (b) transformed right image; (c) the disparity map with respect to (a) and (b); (d) the disparity map with respect to the original left and right images.  }
		\label{fig.PT_DE}
	\end{center}
\end{figure}


In our previous work \cite{Fan2018}, PT-SRP, an efficient subpixel dense stereo algorithm was proposed to reconstruct the 3D road geometry. Although the achieved 3D geometry reconstruction accuracy is higher than 3 mm, the propagation strategy employed in this algorithm is not suitable for GPU programming \cite{fan2019real}. In \cite{fan2019real}, we proposed PT-FBS, a GPU-friendly disparity estimation algorithm, which has been proven to be a good solution to the energy minimization problem in fully connected MRF models \cite{Mozerov2015}. However, its cost aggregation process is a very computationally intensive. Therefore, in this paper, we use SGM \cite{Hirschmuller2008} together with our generalized PT algorithm for 3D road information acquisition. 

\begin{figure}[!t]
	\begin{center}
		\centering
		\includegraphics[width=0.47\textwidth]{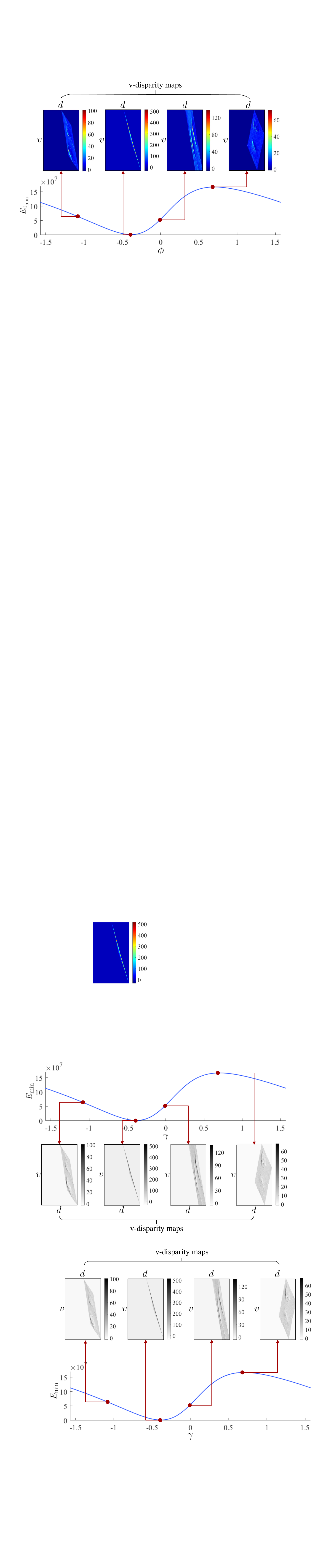}
		\centering
		\caption{$E_{0_\text{min}}$ with respect to different $\phi$.}
		\label{fig.E0_min_phi}
	\end{center}
\end{figure}
In SGM \cite{Hirschmuller2008}, the process of disparity estimation is formulated as an energy minimization problem as follows:
\begin{equation}
\begin{split}
E_1(d_{\mathbf{p}})&=\sum_{\mathbf{p}}\bigg(c(\mathbf{p},d_{\mathbf{p}})+\sum_{\mathbf{q}\in\mathcal{N}_\mathbf{p}}\lambda_1\delta(|d_{\mathbf{p}}-d_{\mathbf{q}}|=1)\\
&+\sum_{\mathbf{q}\in\mathcal{N}_\mathbf{p}}\lambda_2\delta(|d_{\mathbf{p}}-d_{\mathbf{q}}|>1)\bigg),
\end{split}
\label{eq.sgm_energy}
\end{equation}
where $c$ denotes stereo matching cost, $\mathbf{q}$ represents a pixel in the neighborhood system $\mathcal{N}_\mathbf{p}$ of $\mathbf{p}$. $d_\mathbf{p}$ and $d_\mathbf{q}$ are the disparities of $\mathbf{p}$ and $\mathbf{q}$, respectively. $\lambda_1$ penalizes the neighboring pixels with small disparity differences, \ie, one pixel; $\lambda_2$ penalizes the neighboring pixels with large disparity differences, \ie, larger than one pixel. $\delta(\cdot)$ returns 1 if its argument is true and 0 otherwise. However, (\ref{eq.sgm_energy}) is a complex NP-hard problem \cite{Hirschmuller2008}. Therefore, in practical implementation, (\ref{eq.sgm_energy}) is solved by aggregating the stereo matching costs along all  directions in the image using dynamic programming:
\begin{equation}
\begin{split}
&c_\text{agg}^{\mathbf{r}}(\mathbf{p},d_{\mathbf{p}})=c(\mathbf{p},d_{\mathbf{p}})+\min\bigg(c_\text{agg}^{\mathbf{r}}(\mathbf{p}-\mathbf{r},d_{\mathbf{p}}),
\\
&\bigcup_{k\in\{-1,1\}}c_\text{agg}^{\mathbf{r}}(\mathbf{p}-\mathbf{r},d_{\mathbf{p}}+k)+\lambda_1, \min_i c_\text{agg}^{\mathbf{r}}(\mathbf{p}-\mathbf{r},i)+\lambda_2\bigg),
\end{split}
\label{eq.cost_agg}
\end{equation}
where $c_\text{agg}^{\mathbf{r}}(\mathbf{p},d_{\mathbf{p}})$ represents the aggregated stereo matching cost at $\mathbf{p}$ in the direction of $\mathbf{r}$. The disparity at $\mathbf{p}$ can, therefore, be estimated by solving:
\begin{equation}
d_{\mathbf{p}}=\min\sum_{\mathbf{r}}c_\text{agg}^\mathbf{r}(\mathbf{p},d_{\mathbf{p}}).
\end{equation}
The estimated disparity map ${D}_0$ is shown in Fig. \ref{fig.disp0}. Since the right image has been transformed into the left view using PT in Section \ref{sec.PT}, the disparity map ${D}_1$ with respect to the original left and right images, as shown in Fig. \ref{fig.disp1}, can be obtained using: 
\begin{equation}
{D}_1(\mathbf{p})={D}_0(\mathbf{p})	+ \kappa(\mathbf{a}, \mathbf{p}, \phi).
\label{eq.D1}
\end{equation}
\subsection{Disparity Transformation}
\label{sec.disparity_transformation}
\begin{figure}[!t]
	\begin{center}
		\centering
		\subfigure[]
		{
			\includegraphics[width=0.20\textwidth]{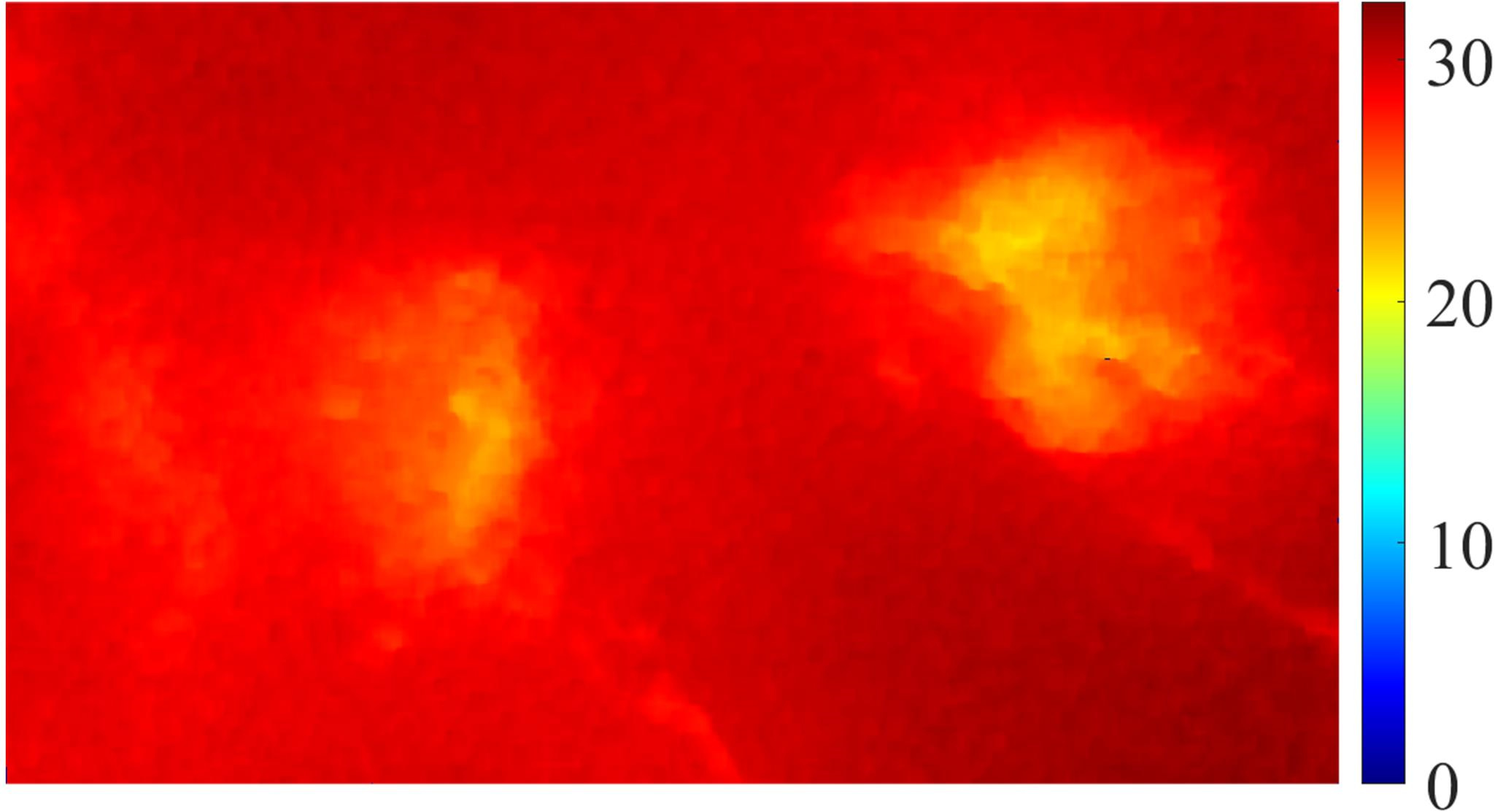}
			\label{fig.disp2}
		}
		\subfigure[]
		{
			\includegraphics[width=0.20\textwidth]{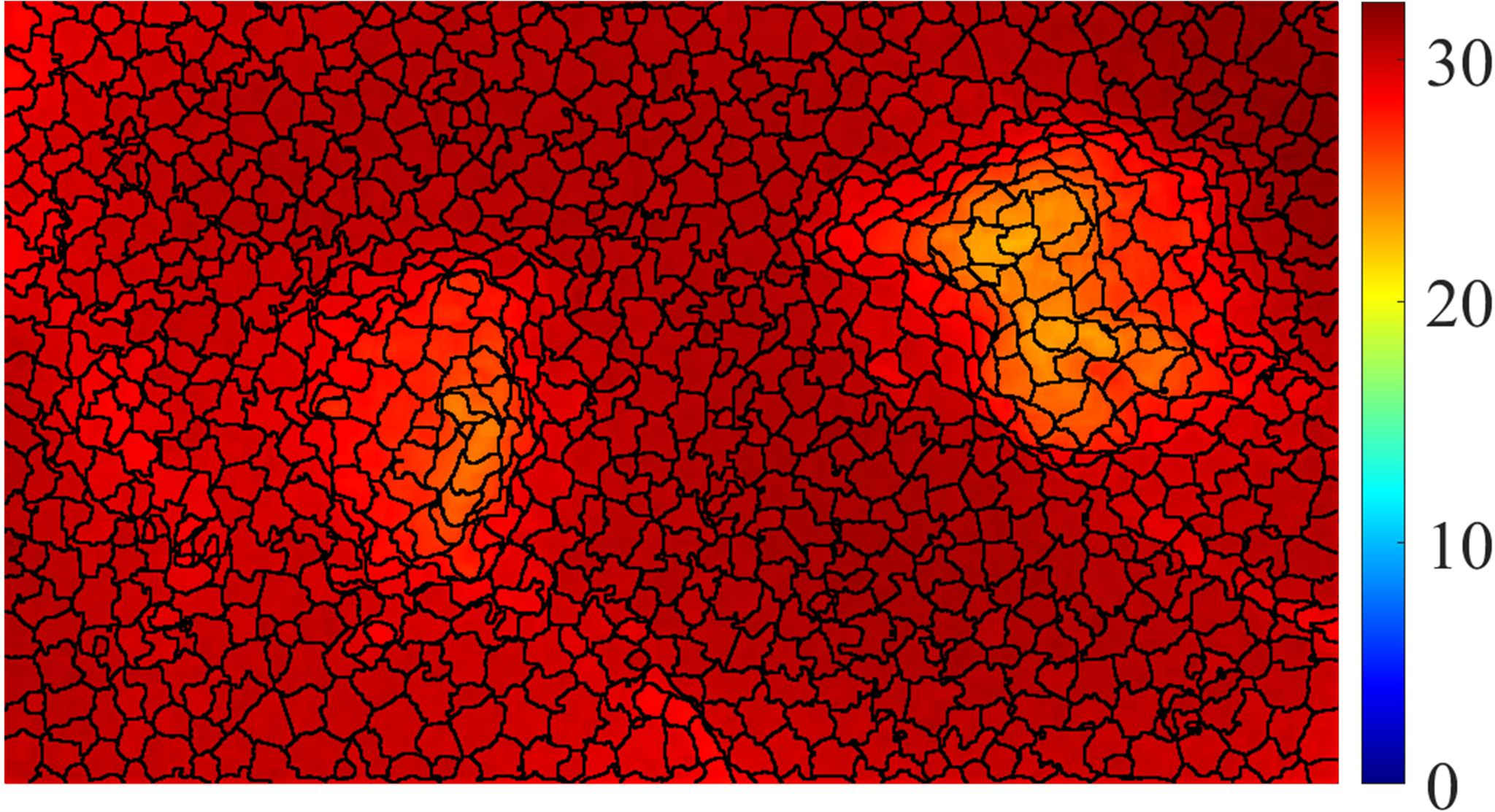}
			\label{fig.superpixel}
		}\\
		\subfigure[]
		{
			\includegraphics[width=0.20\textwidth]{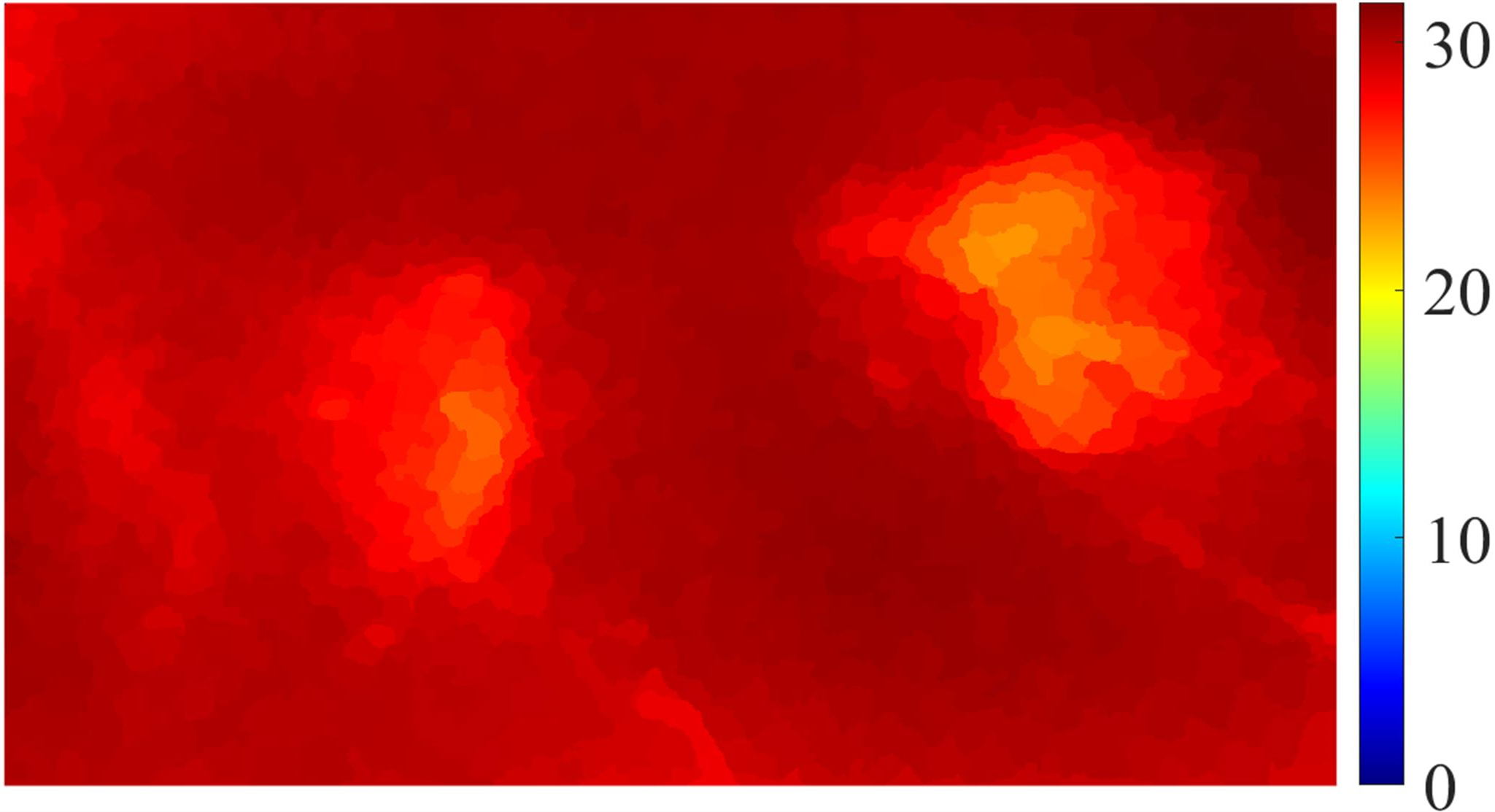}
			\label{fig.disp3}
		}
		\subfigure[]
		{
			\includegraphics[width=0.20\textwidth]{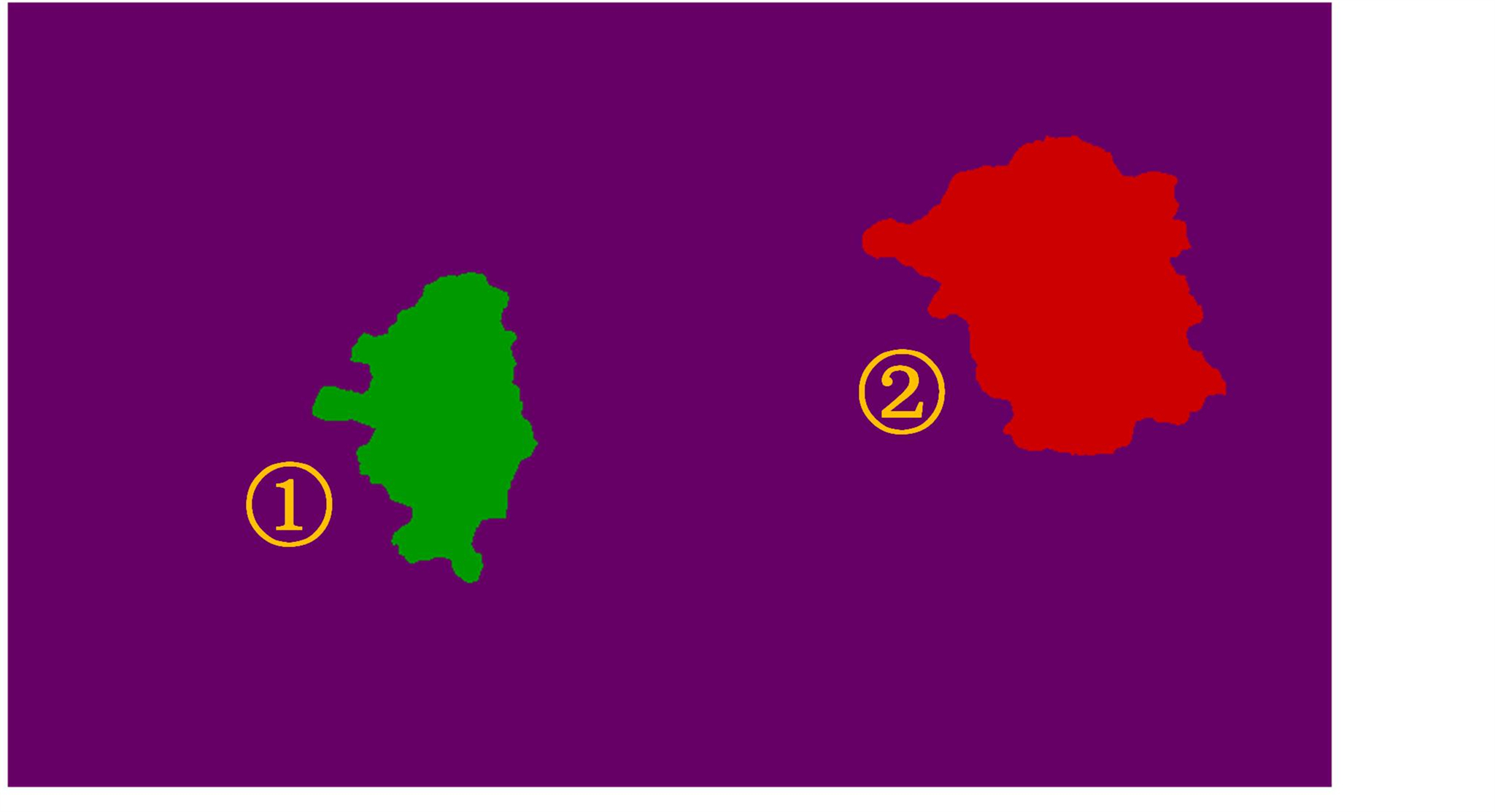}
			\label{fig.detection_res}
		}
		\caption{(a) transformed disparity map; (b) superpixel clustering result; (c) superpixel-clustered transformed disparity map; (d) pothole detection result, where different potholes are shown in different colors.  }
		\label{fig.DT_SS}
	\end{center}
\end{figure}

As discussed in Section \ref{sec.PT}, the road disparity projection can be represented using (\ref{eq.pt}), which has a closed form solution as follows \cite{fan2019pothole}:
\begin{equation}
\mathbf{a}(\phi)=\Big(\mathbf{T}(\phi)^\top\mathbf{T}(\phi)\Big)^{-1}\mathbf{T}(\phi)^{\top}\mathbf{d},
\label{eq.a}
\end{equation}
$E_{0{_\text{min}}}$, the minimum $E_{0}$, has the following expression:
\begin{equation}
\begin{split}
E_{0{_\text{min}}}(\phi)
=\mathbf{d}^\top\mathbf{d}-\mathbf{d}^\top\mathbf{T}(\phi)\Big(\mathbf{T}(\phi)^\top\mathbf{T}(\phi)\Big)^{-1}\mathbf{T}(\phi)^\top\mathbf{d}.
\end{split}
\label{eq.e_min}
\end{equation}
Compared to the case when $\phi=0$, a non-zero $\phi$ results in a much higher $E_{0{_\text{min}}}$ \cite{fan2019real}, as shown in Fig. \ref{fig.E0_min_phi}. 
Therefore, $\phi$ can be obtained by minimizing (\ref{eq.e_min}), which is equivalent to  solving ${\partial E_{0{_\text{min}}}}/{\partial \phi}=0$ and finding its minima \cite{fan2019road}. The solution is given in \cite{fan2019road}. The disparity transformation can then be realized using \cite{fan2020we}: 
\begin{equation}
{D}_2(\mathbf{p})={D}_1(\mathbf{p})	-f(\mathbf{a},\mathbf{p},\phi)+\delta_\text{DT},
\label{eq.D2}
\end{equation}
where ${D}_2$ denotes the transformed disparity map and $\delta_\text{DT}$ is a constant set to ensure that the transformed disparity values are non-negative. The transformed disparity map is shown in Fig. \ref{fig.disp2}, where it can be clearly seen that the damaged road areas become highly distinguishable. 

\subsection{Superpixel Clustering}
\label{sec.superpixel}
In Section \ref{sec.disparity_transformation}, the disparity transformation algorithm allows better discrimination between damaged and undamaged road areas. The road potholes can, therefore, be detected by applying image segmentation algorithms on the transformed disparity maps. However, the  thresholds chosen in these algorithms may not be the best for the optimal pothole detection accuracy, especially when the transformed disparity histogram no longer exhibits an obvious bi-modality.  Furthermore, small blobs with low transformed disparity values are always mistaken for potholes, as discussed in \cite{fan2019road}.  Hence, in this paper, we utilize superpixel clustering to group the transformed disparities into a set of perceptually meaningful regions, which are then used to replace the rigid  pixel grid structure. 

Superpixel generation algorithms are broadly classified as graph-based \cite{Veksler2010, Shi2000} and gradient ascent-based \cite{Vincent1991, Levinshtein2009, Comaniciu2002}. The former treats each pixel as a node and produces superpixels by minimizing a cost function using global optimization approaches, while the latter starts from a collection of initially clustered pixels and refines the clusters iteratively until error convergence. 
In 2012, SLIC \cite{Achanta2012}, an efficient superpixel algorithm was introduced. It outperforms all other state-of-the-art superpixel clustering algorithms, in terms of both boundary adherence and clustering speed \cite{Achanta2012}. Therefore, it is used for transformed disparity map segmentation in our proposed pothole detection system. 

\begin{figure}[!t]
	\begin{center}
		\centering
		\includegraphics[width=0.38\textwidth]{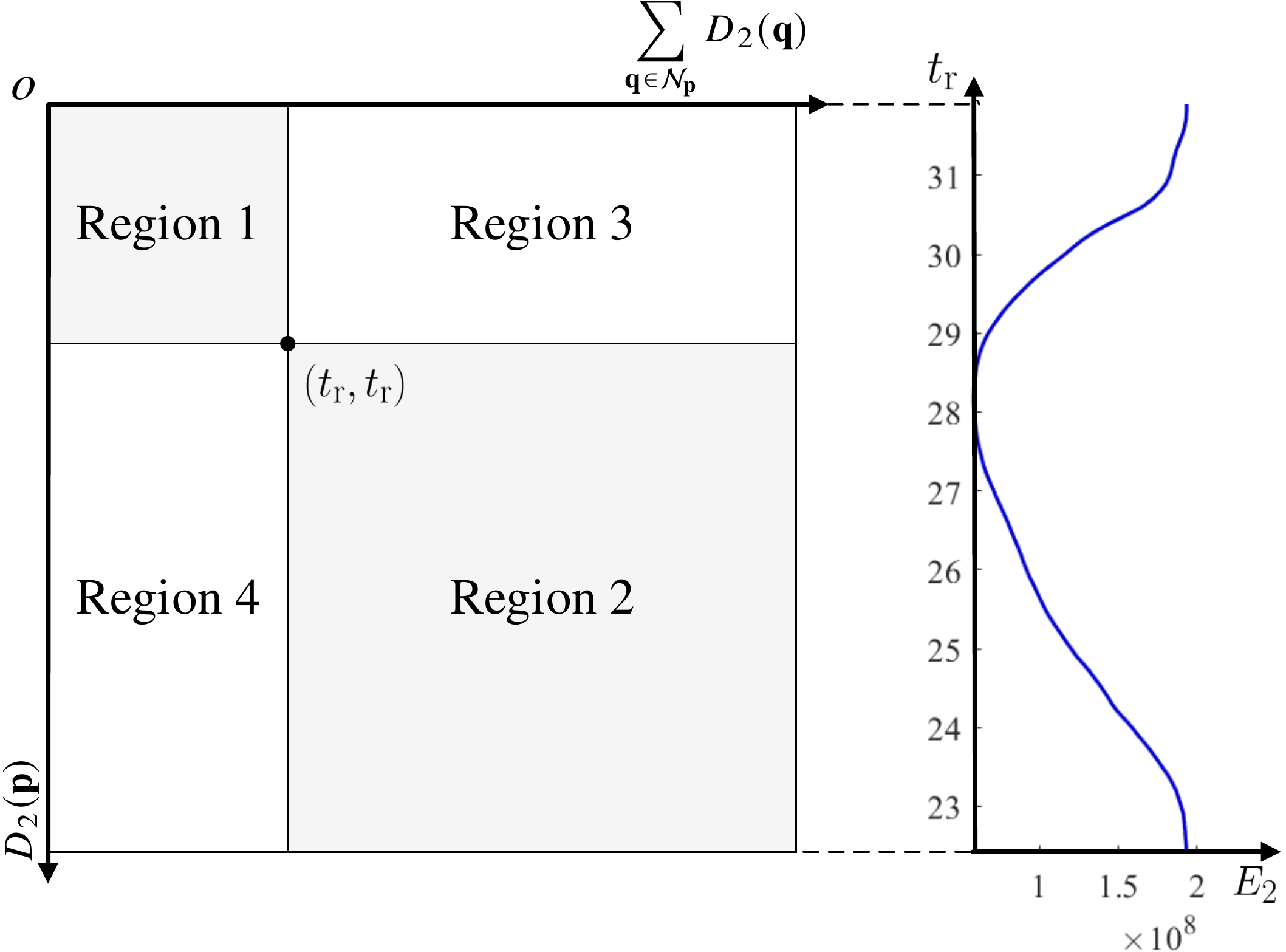}
		\centering
		\caption{An illustration of 2D histogram and $E_2$ with respect to $t_\text{r}$.}
		\label{fig.thresholding}
	\end{center}
\end{figure}
SLIC \cite{Achanta2012} begins with an initialization step, where $p$ cluster centers are sampled on a regular grid to produce $m$ initial superpixels, where $m=\sqrt{HW/p}$. These cluster centers are then moved to the positions (over their eight-connected neighborhoods) corresponding to the lowest gradients. This not only  reduces the chance of selecting a superpixel with a noisy pixel, but also avoids  centering a superpixel on an edge \cite{Achanta2012}. In the next step, each pixel is assigned  to the nearest cluster center, whose search range overlaps its location. Finally, we utilize $k$-means clustering to iteratively update each cluster center   until the residual error  between the previous and updated cluster centers converges. The corresponding SLIC \cite{Achanta2012} result is shown in Fig. \ref{fig.superpixel}, where we can see that each pothole consists of  a group of superpixels.

\subsection{Pothole Detection}
\label{sec.pothole_detection}
After SLIC \cite{Achanta2012}, the transformed disparity map is grouped into a set of superpixels, each of which consists of a collection of transformed disparities with similar values. The value of each superpixel is then replaced by the mean value of its containing transformed disparities, and a superpixel-clustered transformed disparity map $D_3$, as illustrated in Fig. \ref{fig.disp3}, is obtained.  Pothole detection can, therefore, be straightforwardly realized by finding a threshold $t_\text{s}$ and selecting the superpixels whose values are lower than $t_\text{s}$. In this paper, we introduce a novel thresholding method based on $k$-means clustering. The proposed thresholding method hypothesizes that a transformed disparity map only contains two parts: foreground (pothole) and background (road surface), which can be separated using a threshold $t_\text{r}$. Then, we compute the mean value of the transformed disparities in the road surface area. The threshold $t_\text{PD}$ for selecting the pothole superpixels  is then determined as follows:
\begin{equation}
t_\text{s}=t_\text{r}-\delta_\text{PD},
\end{equation}
where $\delta_\text{PD}$ is a tolerance. In this paper, we utilized the brute-force search method to find the best  $\delta_\text{PD}$, which is 2.36. Namely, we select different $\delta_\text{PD}$ between 2 and 8 and compute the overall pixel-level accuracy and F-score. The best  $\delta_\text{PD}$ corresponds to the highest accuracy and F-score.

\begin{figure}[!t]
	\begin{center}
		\centering
		\includegraphics[width=0.40\textwidth]{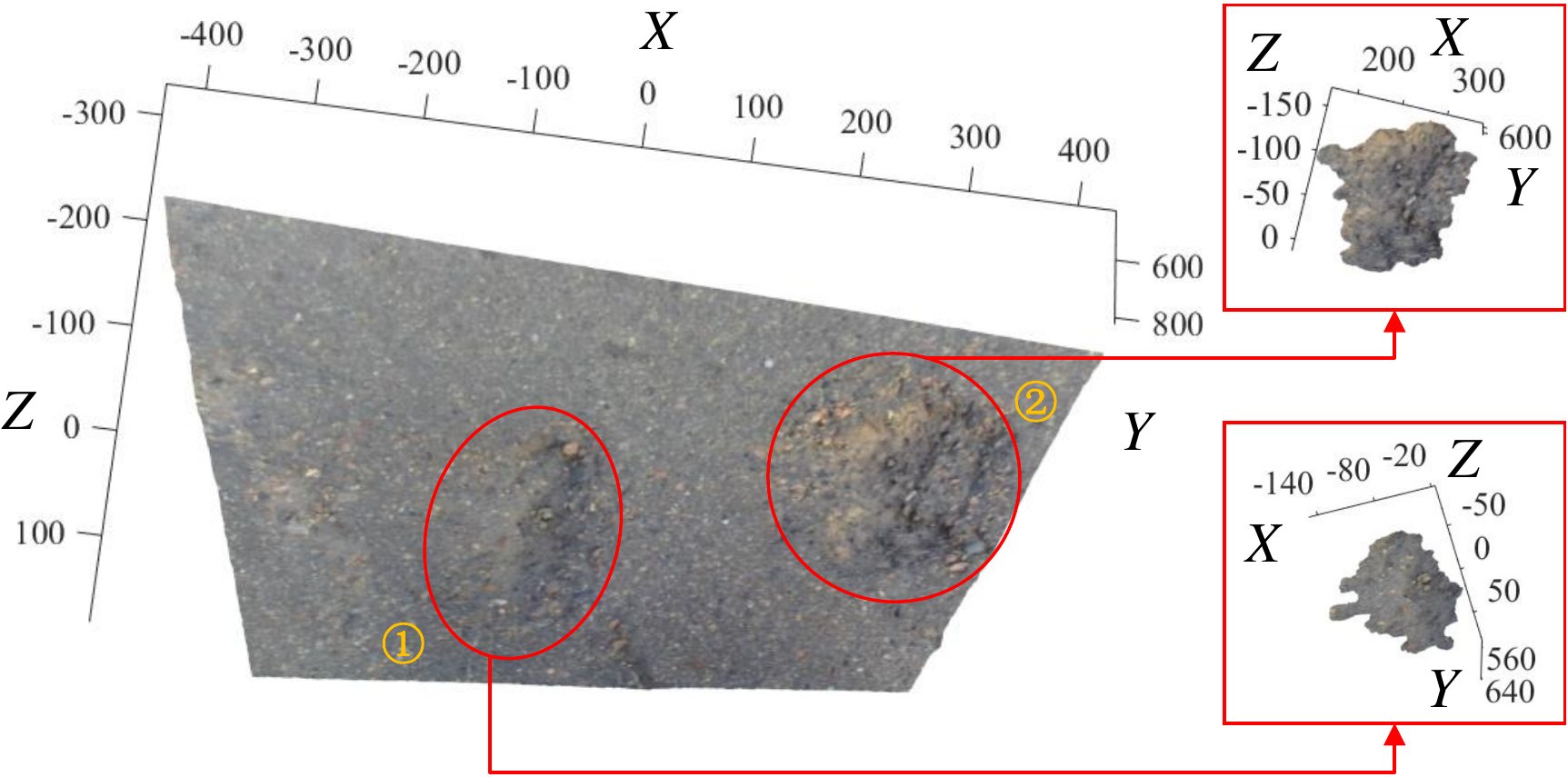}
		\centering
		\caption{The point clouds of the detected potholes.}
		\label{fig.detection_res_3D}
	\end{center}
\end{figure}
To find the best $t_\text{r}$ value, we formulate the thresholding problem as a 2D vector quantization problem, where each transformed disparity ${D}_2(\mathbf{p})$ and its eight-connected neighborhood system $\mathcal{N}_\mathbf{p}$ provide
a vector:
\begin{equation}
\mathbf{g}=[{D}_2(\mathbf{p}),\sum\limits_{\mathbf{q}\in\mathcal{N}_\mathbf{p}}{D}_2(\mathbf{q})]^\top.
\end{equation}
The threshold is determined by partitioning the vectors into two clusters $\mathbf{S}=\{\mathbf{S}_1, \mathbf{S}_2\}$. The vectors $\mathbf{g}$ are stored in a 2D histogram, as shown in Fig. \ref{fig.thresholding}. According to the MRF  theory \cite{Boykov2001}, for an arbitrary point (except for the discontinuities), its transformed disparity value is similar to those of its neighbors in all directions. Therefore, we search for the threshold along the principal diagonal of the 2D histogram, using $k$-means clustering. Given a threshold ${t_\text{r}}$, the 2D histogram can be divided into four regions (see Fig. \ref{fig.thresholding}): regions 1 and 2 represent the foreground and the background, respectively; while regions 3 and 4 store the vectors of noisy points and discontinuities. In the proposed algorithm, the vectors in regions 3 and 4 are not considered in the clustering process. The best threshold is determined by minimizing the within-cluster disparity dispersion, as follows \cite{fan2019crack}:
\begin{equation}
\begin{split}\argminA_{\mathbf{S}}E_2=
\argminA_{\mathbf{S}}\sum_{i=1}^{2}\sum_{\mathbf{g}\in\mathbf{S}_i}||\mathbf{g}-\boldsymbol{\mu}_i||^2,\\
\end{split}
\label{eq.k_mean}
\end{equation}
where $\boldsymbol{\mu}_i$ denotes the mean of points in $\mathbf{S}_i$.  (\ref{eq.k_mean}) can be rearranged as follows:
\begin{equation}
\argminA_{\mathbf{S}}E_2=
\argminA_{\mathbf{S}} \sum_{i=1}^2\sum_{\mathbf{g}\in\mathbf{S}_i}||\mathbf{g}||^2-n_{1}||\boldsymbol{\mu}_1||^2-n_{2}||\boldsymbol{\mu}_2||^2,
\label{eq.k_mean1}
\end{equation}
where $\boldsymbol{\mu}_1$ and $\boldsymbol{\mu}_2$ represent the transformed disparity means in regions 1 and 2, respectively, and $n_{1}$, $n_{2}$ denote the numbers of points in regions 1 and 2, respectively. $E_2$ to ${t_\text{r}}$ is shown in Fig. \ref{fig.thresholding}. The corresponding pothole detection result is shown in Fig. \ref{fig.detection_res}, where different potholes are labeled in different colors using CCL. Also, the point clouds of the detected potholes are extracted from the 3D road point cloud, as shown in Fig. \ref{fig.detection_res_3D}. 
\begin{figure*}[!t]
	\begin{center}
		\centering
		\includegraphics[width=0.99\textwidth]{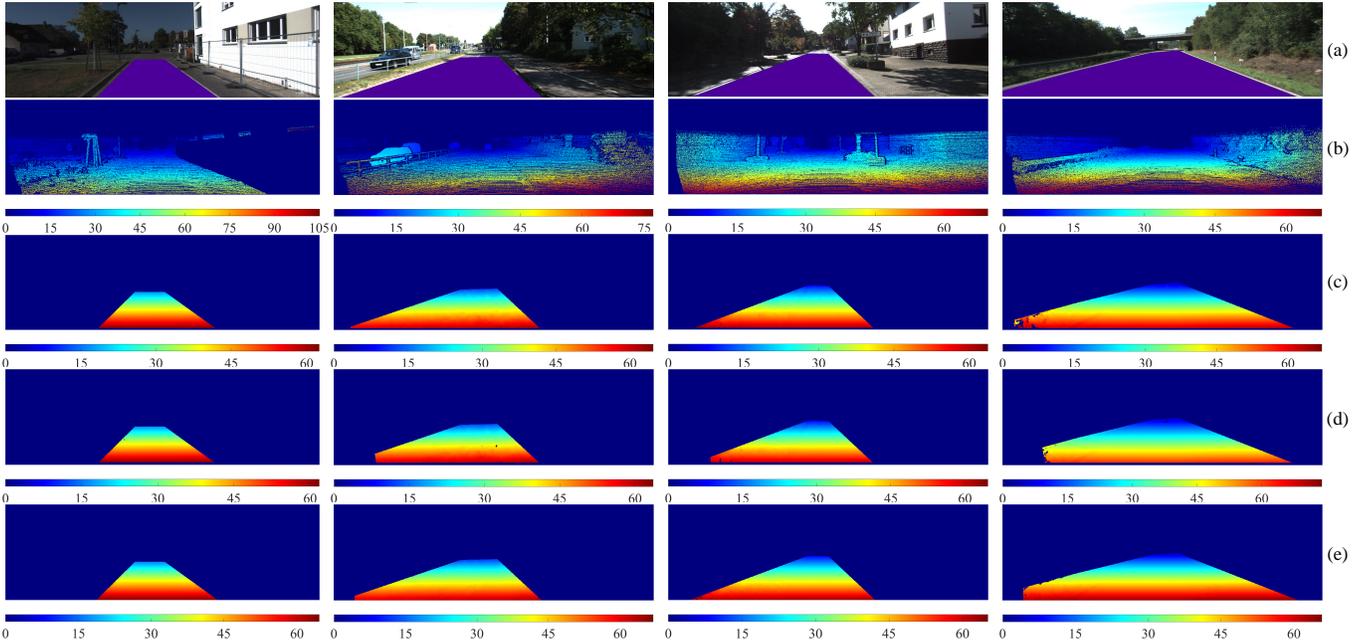}
		\centering
		\caption{Some examples of the disparity estimation experimental results: (a) left stereo images, where the areas in purple are our manually selected road regions; (b) ground truth disparity maps; c)-e) disparity maps estimated using PT-SRP \cite{Fan2018}, PT-FBS \cite{fan2019real} and GPT-SGM, respectively. }
		\label{fig.disp_results}
	\end{center}
\end{figure*}
\section{Experimental Results}
\label{sec.exp}

\begin{figure}[!t]
	\begin{center}
		\centering
		\includegraphics[width=0.48\textwidth]{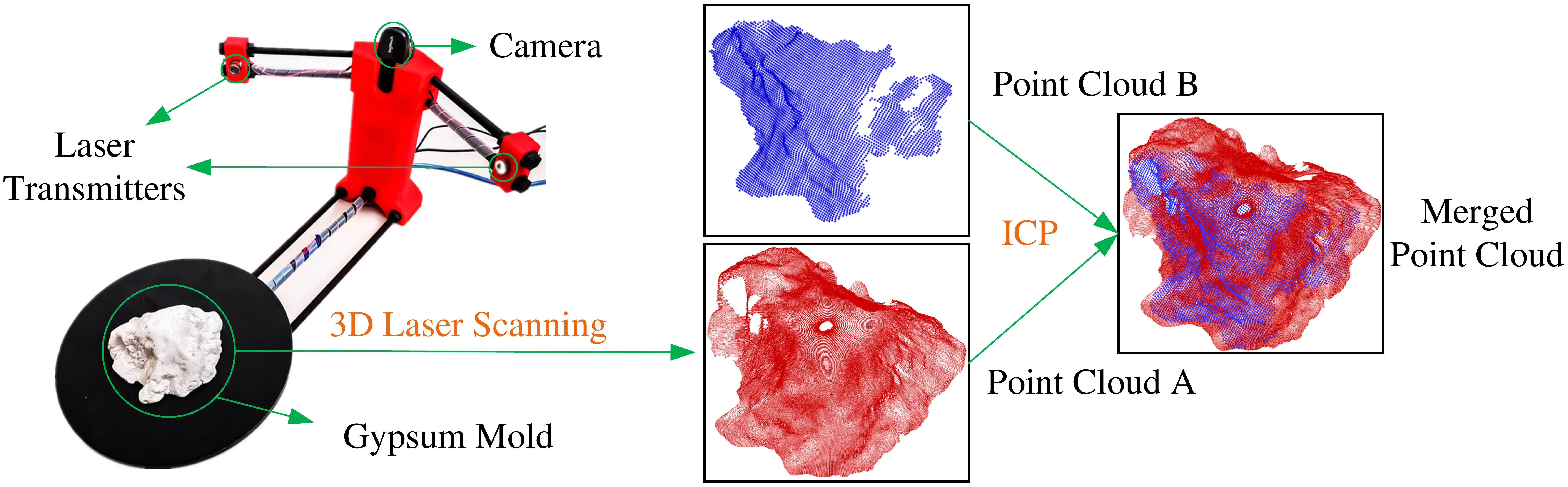}
		\centering
		\caption{Evaluation of 3D pothole reconstruction. Point cloud A is acquired using a BQ Ciclop 3D laser scanner; pothole could B is generated from a dense disparity map. }
		\label{fig.laser_scanner}
	\end{center}
\end{figure}

\begin{figure*}[!t]
	\begin{center}
		\centering
		\includegraphics[width=0.99\textwidth]{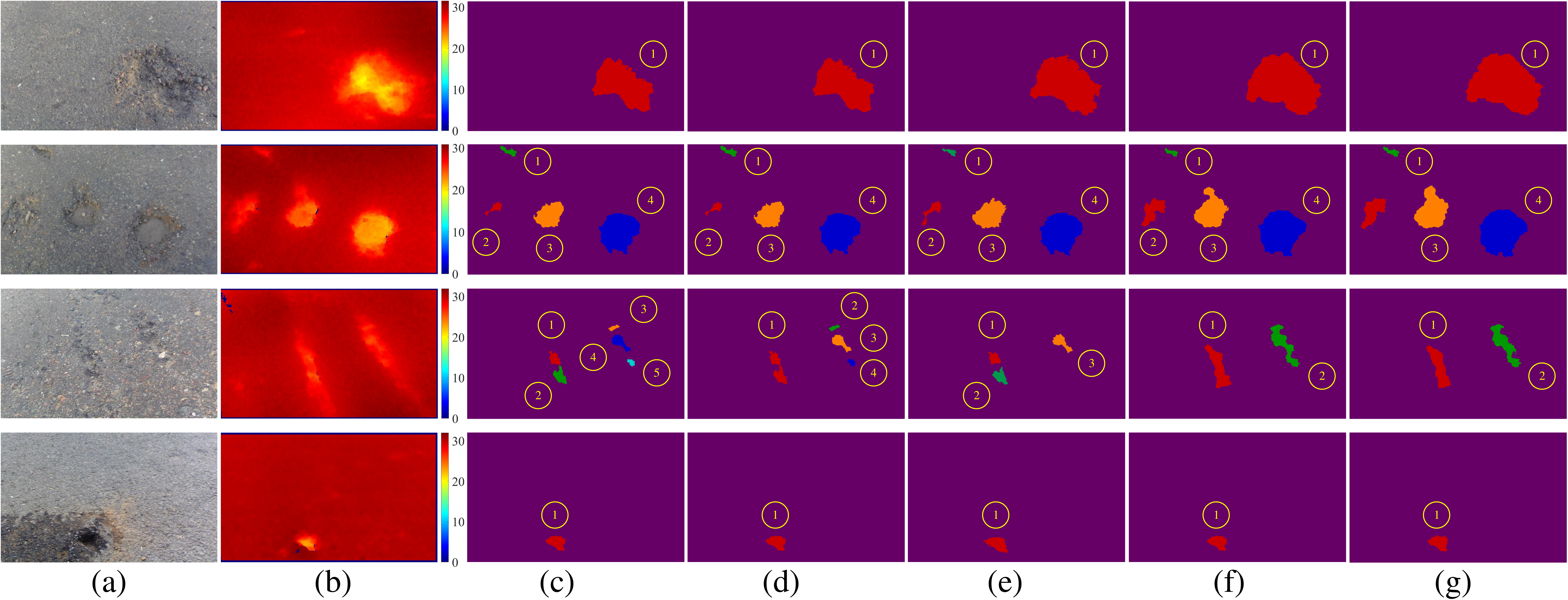}
		\centering
		\caption{Experimental results of pothole detection; a) the left images; b) the transformed disparity maps; c) the results obtained using \cite{Zhang2014}; d) the results obtained using \cite{Mikhailiuk2016}; 3) the results obtained using \cite{fan2019pothole}; f) the results obtained using the proposed algorithm; g) the ground truth.  }
		\label{fig.pd_est_res}
	\end{center}
\end{figure*}
\begin{figure}[!t]
	\begin{center}
		\centering
		\subfigure[]
		{
			\includegraphics[width=0.185\textwidth]{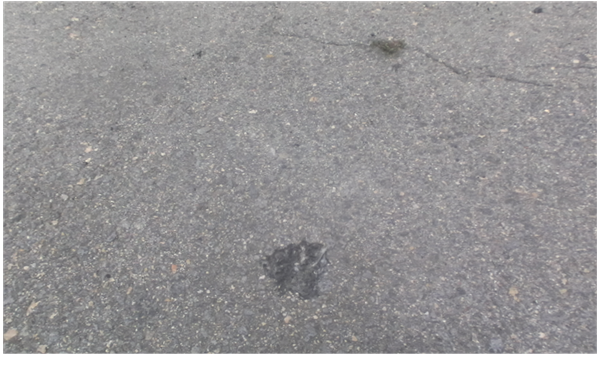}
			\label{fig.f_left}
		}
		\subfigure[]
		{
			\includegraphics[width=0.208\textwidth]{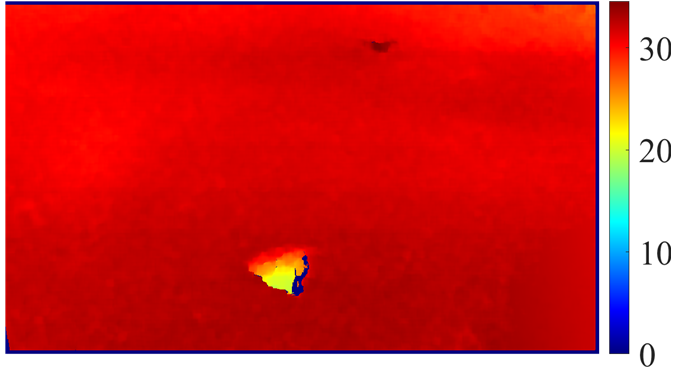}
			\label{fig.f_trans}
		}\\
		\subfigure[]
		{
			\includegraphics[width=0.185\textwidth]{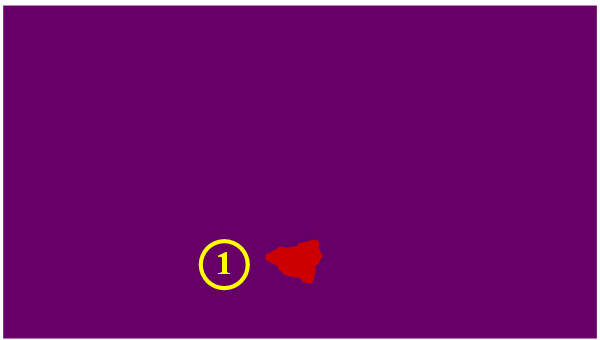}
			\label{fig.f_gt}
		}
		\subfigure[]
		{
			\includegraphics[width=0.208\textwidth]{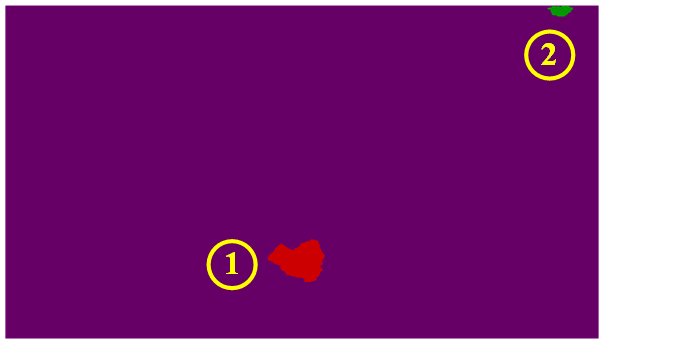}
			\label{fig.f_detection}
		}
		\caption{Incorrect detection:  (a) left image; (b) transformed disparity maps; (c) pothole ground truth; (d) pothole detection result.  }
		\label{fig.incorrect_detection}
	\end{center}
\end{figure}
In this section, we evaluate the performance of our proposed pothole detection algorithm both qualitatively and quantitatively. The proposed algorithm was implemented in CUDA on an NVIDIA RTX 2080 Ti GPU.  The following two subsections discuss the performances of road surface 3D reconstruction and road pothole detection, respectively.



\subsection{Road Surface 3D Reconstruction Evaluation}
\label{sec.eva_disparity_estimation}
In our experiments, we utilized a stereo camera to capture synchronized stereo road image pairs. Our road pothole detection datasets are publicly available at: \url{sites.google.com/view/tcyb-rpd}. 
\subsubsection{Dense Stereo Evaluation}
\label{sec.disp_acu_eva}
Since the road pothole detection datasets we created do not contain disparity ground truth, the KITTI stereo 2012 \cite{Geiger2012} and stereo 2015 \cite{Menze2015} datasets  are used to evaluate the accuracy of our proposed stereo matching algorithm (abbreviated as GPT-SGM).  As our proposed algorithm only aims at estimating the road disparities, we manually selected a road region (see the purple areas in the first row) in each road image to evaluate disparity estimation accuracy.

\begin{table}[!t]
	{	{\begin{center}
				\caption{Comparison of $e_\text{p}$ and $e_\text{r}$ among PT-SRP \cite{Fan2018}, PT-FBS \cite{fan2019real} and our proposed dense stereo algorithm.}
				\label{tab.disp_estimation}
				\begin{tabular}{l|ccc|c}
					\toprule
					\multirow{2}{*}{Algorithm} & \multicolumn{3}{c|}{$e_\text{p}\ (\%)$} & \multirow{2}{*}{$e_\text{r}$}\\
					\cline{2-4}
					& $\varepsilon=1$ & $\varepsilon=2$ & $\varepsilon=3$ \\
					\cline{2-5}
					\hline
					PT-SRP \cite{Fan2018} & 5.0143 & 0.3913  & 0.0588 & 0.4237\\
					PT-FBS \cite{fan2019real} & \textbf{4.5979} & 0.2174 & 0.0227& 0.4092\\
					Proposed & 4.6069 & \textbf{0.1859} & \textbf{0.0083} & \textbf{0.4079}\\
					\bottomrule	
				\end{tabular}
	\end{center}}}
\end{table}
Two metrics are used to measure disparity estimation accuracy :
\begin{itemize}
	\item percentage of error pixels (PEP) \cite{fan2020computer}:
	\begin{equation}
	e_\text{p}=\frac{1}{q}\sum_{\mathbf{p}}\delta\Big(  
	|{D}_1(\mathbf{p})-{D}_4(\mathbf{p}))|>
	\varepsilon\Big)\times100\%,
	\label{eq.e_pep}
	\end{equation}
	where $q$ denotes the total number of disparities used for accuracy evaluation, $\varepsilon$ denotes the disparity error tolerance, and ${D}_4$ represents the ground truth-disparity map. 
	\item root mean squared error (RMSE) \cite{Scharstein2002}:
	\begin{equation}
	e_\text{r}=\sqrt{\frac{1}{q}\sum_{\mathbf{p}}\big(  
		{D}_1(\mathbf{p})-{D}_4(\mathbf{p})
		\big)^2}.
	\label{eq.e_rms}
	\end{equation}
\end{itemize}
Furthermore, we also compare our algorithm with PT-SRP \cite{Fan2018} and PT-FBS \cite{fan2019real}. The experimental results are given in Fig. \ref{fig.disp_results}. Comparisons among these algorithms in terms of $e_\text{p}$ and $e_\text{r}$ are shown in Table. \ref{tab.disp_estimation}, where we can observe that GPT-SGM outperforms PT-SRP \cite{Fan2018} and PT-FBS \cite{fan2019real} in terms of $e_\text{p}$ when $\varepsilon=2$ or $\varepsilon=3$, while PT-FBS \cite{fan2019real} performs slight better than GPT-SGM when $\varepsilon=1$. Furthermore, compared with PT-FBS \cite{fan2019real}, the value of $e_\text{p}$ obtained using GPT-SGM reduces by $14.5\%$ ($\varepsilon=2$) and $63.4\%$ ($\varepsilon=3$), respectively. Moreover, compared with PT-FBS \cite{fan2019real}, when $\varepsilon=1$, $e_\text{p}$ obtained using GPT-SGM increases by only $0.2\%$. Additionally, GPT-SGM achieves the lowest $e_\text{r}$ value (approximately 0.4079 pixels). Therefore, the overall performance of GPT-SGM is better than both PT-SRP \cite{Fan2018} and PT-FBS \cite{fan2019real}. Furthermore, the runtime of our proposed GPT-SGM algorithm on the NVIDIA RTX 2080 Ti GPU is only about 10.1 ms.


\subsubsection{3D Road Geometry Reconstruction Evaluation}
 To acquire the pothole point cloud ground truth, we first poured enough gypsum plaster into a pothole and dug  the gypsum mold out, when it became dry and hardened. 
Then, the 3D pothole model was acquired using a BQ Ciclop 3D laser scanner.
The laser scanner is equipped with a Logitech C270 HD camera and two one-line laser transmitters. The camera captured the reflected laser pulses from the gypsum mold and constructed its 3D model using the calibration parameters. An example of the BQ Ciclop 3D laser scanner and the created pothole ground truth is shown in Fig. \ref{fig.laser_scanner}. Next, we utilized the iterative closest point (ICP) algorithm \cite{Besl1992} to register point clouds A and B, which are acquired using laser scanning and computer stereo vision, respectively. In order to improve the performance of the ICP algorithm, we first transformed the road surface point cloud to make it as close as possible to the $XZ$ plane. This transformation can be straightforwardly realized using the camera height, roll and pitch angles. The merged pothole point cloud is shown in Fig. \ref{fig.laser_scanner}. To quantify the accuracy of  pothole 3D geometry reconstruction, we measure the root mean squared closest distance error $e_\text{c}$:
\begin{equation}
e_\text{c}=\sqrt{\frac{1}{q}\sum_{i=1}^{q}\norm{  
		{\mathbf{P}_{\text{t}_i}}-{\mathbf{P}_{\text{g}_i}}
	}_2^2},
\label{eq.e_c}
\end{equation}
where $\mathbf{P}_{\text{t}}$ denotes a 3D point in the transformed pothole point cloud; $\mathbf{P}_{\text{g}}$ denotes the closest point to $\mathbf{P}_{\text{t}}$ in the ground truth; and $q$ denotes the total number of points used for evaluation. The average $e_\text{c}$ value we achieved is $2.23$ mm, which is lower than what we achieved in \cite{Fan2018}.

\subsection{Road Pothole Detection Evaluation}
\label{seC.eva_pothole_detection}
Some examples of the detected potholes are shown in Fig. \ref{fig.pd_est_res}. In our experiments, the potholes that are either located at the corner of an image or composed of only one superpixel are considered as the fake ones. To evaluate the performance of the proposed pothole detection algorithm, we first compare the detection accuracy of the proposed method with those of the algorithms in \cite{Zhang2014}, \cite{Mikhailiuk2016} and \cite{fan2019pothole}. The results obtained using \cite{Zhang2014}, \cite{Mikhailiuk2016}, and \cite{fan2019pothole} are shown in c), d) and e) of Fig. \ref{fig.pd_est_res}, respectively. The successful detection rates with respect to different algorithms and datasets  are given in Table \ref{table.successful_detection_rate}, where we can see that 
the rates of \cite{Zhang2014} and \cite{Mikhailiuk2016} are $73.4\%$ and $84.8\%$, respectively. The proposed algorithm can detect potholes with a better precision ($98.7\%$), as only one pothole is  incorrectly detected, as  shown in Fig. \ref{fig.incorrect_detection}. The incorrect detection occurs because the road surface at the corner of the image has a higher curvature. We believe this can be avoided by reducing the view angle.

\begin{table*}[!t]
	\begin{center}
		\vspace{0in}
		\footnotesize
		\caption{Comparison of the pothole detection accuracy among \cite{Zhang2014}, \cite{Mikhailiuk2016}, \cite{fan2019pothole} and the proposed one. }
		\label{table.successful_detection_rate}
		\begin{tabular}{c|c|c|c|c|c|c|c|c|c}
			\hline
			Dataset & Method  &\thead{Correct\\Detection} & \thead{Incorrect\\Detection} & Misdetection & Recall & Precision & Accuracy & F-score & Runtime (ms)\\
			\hline
			\multirow{3}{*}{Dataset 1} & \cite{Zhang2014}  & 11 & 11 & 0 & 0.5199 & 0.5427 & 0.9892 & 0.5311 & 33.19\\
			& \cite{Mikhailiuk2016} & 22 & 0 & 0 & 0.4622 & \textbf{0.9976} & 0.9936 & 0.6317 & 22.90 \\
			& \cite{fan2019pothole} & 22 & 0 & 0 & 0.4990 & 0.9871 & 0.9940 & 0.6629 & 117.72 \\
			& Proposed  & 21 & 1 & 0 & \textbf{0.7005}  & 0.9641 & \textbf{0.9947} & \textbf{0.8114} & 47.21 \\
			\hline
			\multirow{3}{*}{Dataset 2} & \cite{Zhang2014}  & 42 & 10 & 0 &  {0.9754} & 0.9712 & {0.9987} & {0.9733} & 30.77\\
			& \cite{Mikhailiuk2016} & 40 & 8 & 4 & 0.8736 & \textbf{0.9907} & 0.9968 & 0.9285 & 21.39 \\
			& \cite{fan2019pothole} &51 & 1 & 0 &  \textbf{0.9804} & 0.9797 & \textbf{0.9991} & \textbf{0.9800} & 124.53 \\
			& Proposed  & 52 & 0 & 0 & 0.9500  & 0.8826  & 0.9920 & 0.9150 & 45.32 \\
			\hline
			\multirow{3}{*}{Dataset 3} & \cite{Zhang2014} & 5 &0 & 0 &  0.6119 & 0.7714 & 0.9948 & 0.6825 & 35.72\\
			& \cite{Mikhailiuk2016} & 5 &0 & 0 & 0.5339 & 0.9920 & 0.9957 & 0.6942 & 26.24  \\
			& \cite{fan2019pothole} & 5 & 0 & 0 & 0.5819 & 0.9829 & 0.9961 & 0.7310 & 132.44\\
			& Proposed   & 5 &0 & 0 &  \textbf{0.7017} & \textbf{0.9961}  & \textbf{0.9964}  & \textbf{0.8234} & 49.90 \\
			\hline
			\multirow{3}{*}{Total} & \cite{Zhang2014}  & 58 &21 & 0 & 0.7799 & 0.8220 & 0.9942 & 0.8004 & 33.23\\
			& \cite{Mikhailiuk2016} & 67 &8 & 4 & 0.6948 & \textbf{0.9921} & 0.9954 &  0.8173 & 23.51\\
			& \cite{fan2019pothole} & 78 & 1 & 0 & 0.7709 & 0.9815 & \textbf{0.9964} & 0.8635 & 124.90 \\
			& Proposed  & 78 &1 & 0 &  \textbf{0.8903} & 0.8982  &{0.9961}  &  \textbf{0.8942} & 47.48\\
			\hline
		\end{tabular}
	\end{center}
\end{table*}

\begin{table*}[!t]
	\begin{center}
		\vspace{0in}
		\footnotesize
		\caption{Comparison of four state-of-the-art DCNNs trained for road pothole detection. }
		\label{table.deeplearning}
		\begin{tabular}{l|c|c|c|c|c|c}
			\hline
			{\multirow{2}*{DCNN}} & \multicolumn{2}{c|}{Disp} & \multicolumn{2}{c|}{TDisp} & \multicolumn{2}{c}{RGB}
			\\
			\cline{2-7}
			& accuracy & F-score & accuracy & F-score & accuracy & F-score \\
			\hline
			FCN \cite{long2015fully} & 0.971 & 0.606 & 0.983 & 0.797 &  0.949 &  0.637 \\
			SegNet \cite{badrinarayanan2017segnet} & 0.966 &  0.516 & 0.979 &  0.753 & 0.894 & 0.463 \\
			U-Net \cite{ronneberger2015u} &  $\mathbf{0.971}$ & 0.639 & 0.984 &  0.805 & 0.966 &  0.633\\
			DeepLabv3+ \cite{chen2018encoder} & 0.968 & $\mathbf{0.673}$ & $\mathbf{0.987}$ & $\mathbf{0.856}$ & $\mathbf{0.977}$ & $\mathbf{0.742}$\\
			\hline
		\end{tabular}
	\end{center}
		\vspace{-2em}
\end{table*}
Furthermore, we also compare these pothole detection algorithms with respect to the pixel-level precision, recall, accuracy and F-score, defined as follows: $\text{precision}=\frac{n_\text{tp}}{n_\text{tp}+n_\text{fp}}$, $\text{recall}=\frac{n_\text{tp}}{n_\text{tp}+n_\text{fn}}$, $\text{accuracy}=\frac{n_\text{tp}+n_\text{tn}}{n_\text{tp}+n_\text{tn}+n_\text{fp}+n_\text{fn}}$ and $\text{F-score}=2\times\frac{\text{precision}\times\text{recall}}{\text{precision}+\text{recall}}$, 
where $n_\text{tp}$, $n_\text{fp}$, $n_\text{fn}$ and $n_\text{tn}$ represent the numbers of  true positive, false positive, false negative and true negative pixels, respectively. The comparisons with respect to these four performance metrics are also given in Table \ref{table.successful_detection_rate}, where it can be seen that the proposed algorithm outperforms \cite{Zhang2014}, \cite{Mikhailiuk2016} and \cite{fan2019pothole} in terms of both accuracy and F-score when processing datasets 1 and 3. \cite{fan2019pothole} achieves the best performance on dataset 2.  In addition, our method achieves the highest overall  F-score ($89.42\%$), which is even over $3\%$ higher than our previous work \cite{fan2019pothole}. 

Also, we provide the runtime of \cite{Zhang2014},  \cite{Mikhailiuk2016}, \cite{fan2019pothole} and our proposed method on the NVIDIA RTX 2080 Ti GPU in Table \ref{table.successful_detection_rate}. It can be seen that the proposed system performs much faster than \cite{fan2019pothole}. Although our proposed method performs slower than \cite{Zhang2014} and \cite{Mikhailiuk2016}, SLIC \cite{Achanta2012} takes the biggest proportion of the processing time. The total runtime of DT and pothole detection is only about 3.5 ms. Therefore, we believe by leveraging a more efficient superpixel clustering algorithm the overall performance of our proposed pothole detection system can be significantly improved. Moreover, as discussed above, the proposed pothole detection performs much more accurately than both \cite{Zhang2014} and \cite{Mikhailiuk2016}, where an increase of approximately 9\% is witnessed on the F-score.

Additionally, many recent semantic image segmentation networks have been employed to detect freespace (drivable area) and road pothole/anomaly \cite{fan2020sne, wang2020applying, fan2020we}. Therefore, we also compare the proposed algorithm with four state-of-the-art deep convolutional neural networks (DCNNs): 1) fully connected network (FCN) \cite{long2015fully}, SegNet \cite{badrinarayanan2017segnet}, U-Net \cite{ronneberger2015u} and DeepLabv3+ \cite{chen2018encoder}. Since only a very limited amount of road data are available, we employ $k$-fold cross-validation \cite{geisser1975predictive} to evaluate the performance of each DCNN, where $k$ represents the total number of images. Each DCNN is evaluated $k$ times. Each time, $k-1$ subsamples (disparity maps, transformed disparity maps, or RGB images) are used to train the DCNN, and the remaining subsample is retained for testing DCNN performance. Finally, the obtained $k$ groups of evaluation results are averaged to illustrate the overall performance of the trained DCNN. The quantification results are given in Table \ref{table.deeplearning}, where it can be observed that the DCNNs trained with the transformed disparity maps (abbreviated as TDisp) outperform themselves trained with either disparity maps (abbreviated as Disp) or RGB images (abbreviated as RGB). This demonstrates that DT makes the disparity maps become more informative. Furthermore, Deeplabv3+ \cite{chen2018encoder} outperforms all the other compared DCNNs for pothole detection. However, it can be observed that our proposed pothole detection algorithm outperforms all the compared DCNNs in terms of both accuracy and F-score. We believe this is due to that only a very limited amount of road data are available, and therefore, the advantages of DCNNs cannot be fully exploited.

\section{Discussion}
\label{sec.discussion}

Potholes are typically detected by experienced inspectors in fine-weather daylight, which is an extremely labor-intensive and time-consuming process. The proposed road pothole detection algorithm can perform in real time on a state-of-the-art graphics card. Compared with the existing DCNN-based methods, our algorithm does not require labeled training data to learn a pothole detector. The accuracy we achieved is much higher than the existing computer vision-based pothole detection methods, especially those based on 2D image analysis. Although computer vision-based road pothole detection has been extensively researched over the past decade, only a few researchers have considered to apply computer stereo vision in pothole detection. Therefore, we created three pothole datasets using a stereo camera to contribute to the research and development of automated road pothole detection systems. In our experiments, the stereo camera was mounted to a relatively low height to the road surface, in order to increase the accuracy of disparity estimation. Our datasets provide pixel-level road pothole ground truth as well as pothole 3D geometry ground truth, which can be used by other researchers to quantify the accuracy of their developed road surface 3D reconstruction and road pothole detection algorithms.

\section{Conclusion and Future Work}
\label{sec.con}
In this paper, we presented an efficient stereo vision-based pothole detection system. We first generalized the PT algorithm \cite{Fan2018} by considering the stereo rig roll angle into the process of PT parameter estimation. DT made potholes highly distinguishable from the undamaged road surface. SLIC grouped the transformed disparities into a collection of superpixels. Finally, the potholes were detected by finding the superpixels, which have lower values than an adaptive threshold determined using $k$-means clustering. The proposed pothole detection system was implemented in CUDA on a RTX 2080 Ti GPU. The experimental results illustrated that our system can  achieve a successful detection rate of $98.7\%$ and a pixel-level accuracy of $99.6\%$. 

A challenge is that the road surface cannot always be considered as a ground plane, resulting in wrong detection. Therefore, we plan to design an algorithm to segment the reconstructed road surface into different planar patches, each of which can then be processed separately, using the proposed algorithm. 


\bibliographystyle{IEEEtran}
\end{document}